# DEVELOPING MULTI-MODAL MACHINE LEARNING MODEL FOR PREDICTING PERFORMANCE OF AUTOMOTIVE HOOD FRAMES

**Abhishek Indupally**
Clemson University
Clemson, SC

**Satchit Ramnath***
Clemson University
Clemson, SC

**ABSTRACT**

*Is there a way for a designer to evaluate the performance of a given hood frame geometry without spending significant time on simulation setup? This paper seeks to address this challenge by developing a multimodal machine-learning (MMML) architecture that learns from different modalities of the same data, to predict performance metrics. It also aims to use the MMML architecture to enhance the efficiency of engineering design processes by reducing reliance on computationally expensive simulations in the initial phases of design. The proposed architecture helps accelerate design exploration, enabling rapid iteration while maintaining high accuracy in predictions. The study also presents results that show that by combining multiple data modalities, MMML outperforms traditional single-modality (or unimodal) approaches. Two new frame geometries, not part of the training dataset, are also used for prediction using the trained MMML model to showcase the ability to generalize to unseen frame models. The findings underscore MMML's potential in supplementing traditional simulation-based workflows, particularly in the conceptual design phase, and highlight its role in bridging the gap between machine learning and real-world engineering applications. This research paves the way for the broader adoption of machine learning techniques in engineering design, with a focus on refining multimodal approaches to optimize structural development and accelerate the design cycle.*

## 1. INTRODUCTION

In engineering design, a structured and systematic approach is essential for developing effective solutions [1]. One of the primary tasks in the conceptual design phase is assessing the structural integrity and overall performance of the design. Traditionally, evaluating structural integrity requires computational simulations such as Finite Element Analysis (FEA), which involves defining boundary conditions, running simulations, and validating results [2]. While highly accurate, and also necessary, FEA is computationally expensive and time-intensive, potentially creating bottlenecks in the iterative design process during the initial phases. This slows down innovation and increases the design cycle time, affecting the number of design alternatives explored. Reducing design cycle time is, therefore, crucial, as it enables engineers to assess multiple configurations more efficiently, refine their designs in real time, and accelerate product development without compromising accuracy.

Machine learning (ML) advancements offer to supplement traditional simulation-driven design workflows. ML-based models can learn from existing simulation data and provide fast, approximate performance metrics predictions, allowing engineers to evaluate design iterations in real time. This rapid feedback loop enhances the design process by enabling quick assessments and iterative refinements.

In the context of an automotive hood design, the hood consists of two primary components: the outer skin and the hood frame, as shown in Figure 1. The hood skin refers to the outer, visible panel (class-A surface) of the hood, defining the vehicle's aesthetics and aerodynamics, while the hood frame is the structural support underneath that panel, providing structural integrity, crash safety, and attachment points. The frame consists of various geometrical features, as shown in Figure 2. The ribs are considered the primary features since they provide structural stiffness, and the pockets are the secondary features that help reduce the weight. Since the hood skin dictates the design space and the curvature, the frame must be engineered to follow the same design guidelines while still satisfying performance requirements such as weight and strength.

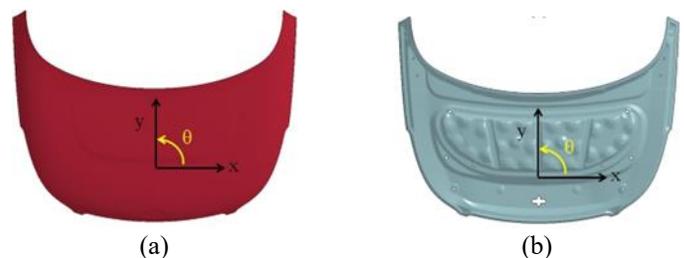

*Figure 1: (a) Automotive hood skin and (b) the frame [3]*

---

* Corresponding author: sramnat@clemson.edu

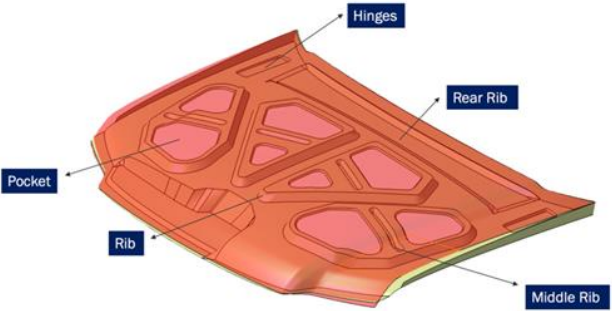

*Figure 2: Primary features on a hood frame (Ribs), secondary features on a hood frame (Pockets)*

A designer developing a hood frame must assess its structural integrity using FEA by predicting key performance metrics, which may include stresses and deflections. However, relying solely on FEA for these evaluations is effort and time-intensive, as each iteration requires re-running simulations, creating a computational bottleneck that limits rapid design exploration. Effective feedback during the design cycle [4] is crucial, as it allows designers to quickly estimate performance metrics, assess structural integrity, and refine their concepts without the time-intensive burden of full-scale simulations. To enable this, a method for quick performance estimation is needed—one that enables designers to analyze localized effects of geometric changes, add or remove structural features, and understand how these modifications impact performance without the computational overhead of full-scale simulations.

Although these methods aid in conceptual automotive body designs, predicting structural performance accurately during the design process requires a model that can effectively capture and interpret complex geometric relationships. Capturing the complex details and relationships may not explicitly be available in a dataset, and hence requires new model architectures to extract, learn, and build the implicit relationships. This can be achieved by extracting data in different representations called "*modalities*". The concept of modality is foundational in multimodal machine learning, referring to a specific representation of information in a particular format. Liang et al. [5] describe modality as "a way in which a natural phenomenon is perceived or expressed". This definition encompasses various types of representations, including visual, auditory, textual, and numerical data. Although research has been done to evaluate various ML models to predict the performance of designs, especially for automotive body design, the input data to these models are unimodal - only one modality - like images or point clouds or parameters.

This research proposes a Multi-Modal ML (MMML) architecture that can process different modalities to make accurate predictions. In this research, the three primary modalities considered are: image data, geometric data, and parametric data. The approach integrates image-based features extracted from convolutional neural networks (CNNs) with geometric and parametric data to improve the overall predictive accuracy. Figure 3 shows the workflow for the MMML model. The images provide spatial and visual information; the geometric data includes cross-sectional profiles of the design, and the parametric data includes parameters corresponding to geometric features in the model. Combining multiple modalities enables a more comprehensive understanding of a design's behavior. The contributions of this research are:

- Development of a Custom MMML Architecture – Design and develop a MMML architecture that integrates various modalities to predict key performance metrics.
- Comparison with Single-Modality Models – The proposed MMML architecture is evaluated against a baseline model trained with a single modality to assess the benefits of multimodal learning.
- Generalization to Unseen Designs – The model is tested on new hood frame designs to evaluate its robustness, predictive accuracy, and ability to generalize beyond the training data.

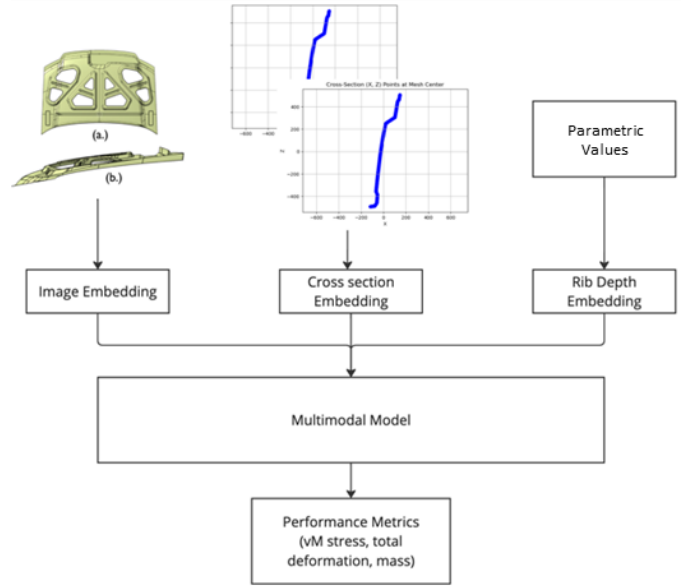

*Figure 3: Multi-modal machine learning flowchart*

## 2. BACKGROUND

Machine learning algorithms are increasingly used in automotive body design to optimize safety, performance, material efficiency, and cost. By analyzing large datasets, these algorithms help predict outcomes and improve decision-making in areas like crashworthiness, panel rigidity, and material usage. ML can also help automate design tasks, speeding up development cycles and aiding innovation.

### 2.1 Machine Learning in Automotive Design

There has been quite a bit of research into integrating ML algorithms into the domain of automotive design. Liu et al. [6] proposed a structured framework to develop design-specific knowledge graphs, which are then used in deep learning to learn graph embeddings, make predictions, and support reasoning. Bodendorf and Franke [7] compared various machine learning approaches to predict the costs in an early product design phase of automotive components. The study looked at the performance of six machine learning algorithms for estimating costs. Felix et al. [8] proposed an approach that mirrors the team structure of engineers, CAD designers, and quality assurance, with a Vision Language Model (VLM)-based Multi-Agent System with access to parametric CAD tooling and documentation. Combining agents for requirements engineering, CAD engineering, and vision-based quality assurance, a model is generated automatically from sketches and/ or textual descriptions. Wong et al. [9] proposed an end-to-end prompt evolution design optimization (PREDO) framework contextualized in a vehicle design scenario that leverages a vision-language model to penalize impractical car designs synthesized by a generative model. The research used an evolutionary strategy coupled with an optimization objective function that comprises a physics-based solver and a vision-language model for practical or functional guidance in the generated car designs. Rios et al. [10] propose and realize a fully automated evolutionary design optimization framework using Shap-E [11] in the context of aerodynamic vehicle optimization. Zhen et al. [12] proposed a method for automotive hood design based on machine learning-assisted structural optimization. The objective of the study was to optimize the hood frame for pedestrian head impact loads.

## 2.2 Multi-Modal Machine Learning

Multimodal machine learning tasks emerge when inputs or outputs are represented in various formats or when different types of atomic units of information are combined within the same system [13]. These models have gained significant traction due to their ability to leverage complementary information from multiple modalities, leading to more informed predictions. For instance, visual information can be enhanced with textual descriptions or numerical features to generate more precise models that account for real-world complexities. Integrating multiple modalities has proven highly effective in improving prediction accuracy and generalization across various machine-learning applications. Baltrusaitis et al. [14] provide a comprehensive overview of multimodal machine learning, highlighting five main technical challenges: representation, translation, alignment, fusion, and co-learning. Their work emphasizes that multimodal models often outperform unimodal counterparts by capturing diverse aspects of input data, thereby mitigating the limitations of any single modality.

Multimodal machine learning (MMML) has demonstrated significant advancements in predictive modeling, particularly in automotive engineering. Su et al. [15] highlight the benefits of integrating multiple data modalities over unimodal approaches, which rely on a single data source and may lead to incomplete insights. The MMML model used in the research combines images, text, and parametric data to predict five key vehicle rating scores, achieving a 4% to 12% improvement over unimodal models. This underscores the value of leveraging diverse data types for more accurate predictions. While the work focuses on vehicle ratings, a similar challenge exists in evaluating the performance of CAD-based mechanical components. Structured parametric data often provides the most informative input, but non-visual data, such as text, can sometimes outperform images in predictive tasks.

In computer-aided design (CAD), particularly in conceptual design, integrating multiple data representations can enhance the design process. Li et al. [16] introduced LLM4CAD, a large language model that combines images, sketches, and textual descriptions to generate 3D geometry. The research underscores the importance of multimodal approaches in CAD, demonstrating how the fusion of different input modalities can lead to more expressive and effective design workflows. By bridging visual and textual representations, MMML facilitates interactive and sophisticated design processes, enabling designers to explore and refine concepts more efficiently. A key takeaway from LLM4CAD is the crucial role of high-quality datasets in the success of multimodal models. As research in this domain progresses, developing diverse, domain-specific datasets will be essential for improving predictive performance and generalizability in engineering design models.

A study by Gong et al. [17] on bimodal learning in materials science highlights the advantages of combining composition and structure data to predict material properties. The framework processes composition as vectors and structure as graphs, fusing these modalities to enhance predictive performance. The results demonstrate that bimodal models consistently outperform unimodal approaches by capturing complementary information, reinforcing the value of integrating diverse data types. The research also highlights the importance of effective data representation for machine learning applications.

The study by Kumar [18] on geometrical parameter extraction from flexible mechanical components and assemblies provides critical insights into the application of machine learning for mechanical performance prediction. By leveraging the CarHoods10K [19] dataset, a comprehensive collection of 10,000 validated automotive hood frame geometries, the study establishes a strong foundation for data-driven design evaluation. One of the key contributions of Kumar's work is the multi-view approach (using images), which demonstrates how various views of hood frames can enhance structural understanding. The method uses images to extract features for performance prediction. However, the study also reveals a fundamental limitation of image-only models—while they effectively capture visual and shape-based features, they lack structured, numerical context that could provide deeper insights into mechanical behavior.

Hwang et al. [20] present EMMA, a new approach to improving autonomous driving systems by using artificial intelligence (AI) to process and understand multiple types of information at once. Traditional self-driving systems rely on separate components for tasks like detecting objects, planning movements, and understanding road layouts. These components work independently and pass information between them, which can sometimes lead to errors and inefficiencies. Instead of using separate modules, EMMA takes in all the necessary data—like camera images, navigation directions, and vehicle status—and processes it in a single unified AI model. This allows the system to make driving decisions more efficiently and accurately. One of the key innovations is the use of Multimodal Large Language Models (MLLMs), a type of AI that can combine different types of inputs (such as images and text) into a single understanding of the driving environment. While EMMA is a promising step toward smarter self-driving AI, the authors acknowledge some limitations, such as high computational requirements and challenges in processing depth and distance information because it does not use LiDAR (a laser-based distance measurement technology).

Acosta et al. [21] explore how integrating multiple data sources—such as medical images, genomics, biosensors, and clinical records—can enhance artificial intelligence (AI) applications in healthcare. The authors highlight the limitations of current AI models, which typically rely on a single data modality and provide only moment-in-time assessments, whereas real-world medical decisions require continuous, multimodal inputs. A key contribution of this work is its discussion of multimodal AI models, which can analyze diverse biomedical data types, including wearable sensors, genetic and epigenetic markers, imaging, clinical text, and social determinants of health. By combining these sources, multimodal AI enables more accurate diagnosis, prognosis, and treatment planning, bridging gaps in current medical AI applications.

The proposed research builds upon these foundations, leveraging multimodal learning to predict the performance of automotive hood frames. By combining image data with geometric and parametric information, the research aims to overcome the limitations of unimodal and multi-view CNN approaches and provide more accurate, robust predictions for structural performance.

## 3. DATASET EXTRACTION AND PREPROCESSING

The CarHoods10K dataset offers over 10,000 3D mesh automobile hood frame geometries developed by an industry-standard CAD generation process [19][22]. The dataset contains highly realistic, expert-validated designs that are created by generating over 100 parameterized shape variants (from skins and feature patterns) through a feature-based modeling approach. This technique enables the creation of diverse designs by adjusting parameters that influence key features like ribs and cut-outs, which are essential for the structural integrity of the hoods. Each model in the dataset is available in STL format, along with design parameters and performance metrics from finite element analysis (FEA), such as stress, deformation, and mass. These details are vital for developing advanced computational and ML models to predict performance in automotive engineering. Figure 4 displays the 100 shape variants in the geometry dataset. A feature-based design approach was used to generate the large dataset, details of which can be found in [23]. The performance data for the models in the dataset, which include vonMises stress, direction deflection, and mass, were obtained from FEA as described in [24]. The proposed method uses the dataset to extract data – images, geometric data, and parametric data – for training and validating advanced architectures that will aid designers in evaluating new designs rapidly.

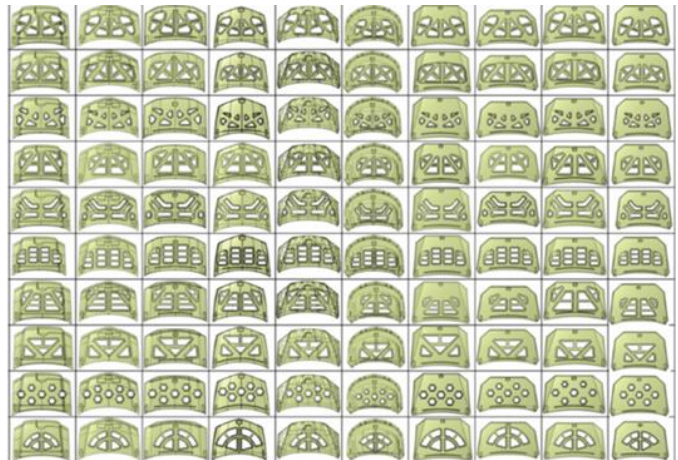

*Figure 4 : CarHoods10K dataset shape variations*

### 3.1 Image Dataset (Top and Side Views)

The top (or normal direction) view of the hood frame is extracted and used to give the ML model context about the type, number, and position of various geometrical features with respect to the design space of the frame, while the side views give context about the curvature and "flow" of the hood frame. By combining these perspectives, the ML model can better understand and extract the object's shape, contents, and structure. These features, extracted from the images, are essential for predicting the performance of the hood frame, such as load-bearing capacity or stress distribution. The images are processed using two separate Residual Networks (ResNet50) [25]. ResNet was chosen because of the efficiency and speed of training the model. ResNet50 makes predictions by categorizing an image, and in this case, the goal is to train the model to predict the frame's performance based on the extracted features. Figure 5 shows an example of top and side views of a hood frame.

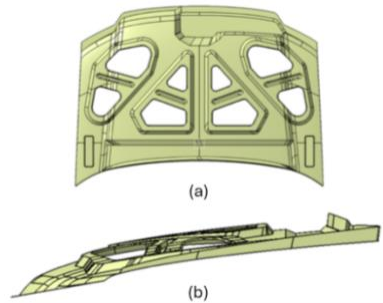

*Figure 5: (a) Top view (b) side view of a hood frame*

### 3.2 Geometric and Parametric Data (Cross-sections & Feature Depths)

Given the change in curvature along the hood frame, extracting the cross-section at two locations – 25% and 75% along the width – would help provide additional information about the frame to the algorithm. This is the geometric data or modality that is part of the MMML algorithm. The change in curvature of the frame is not "visible" or available from the image data. Figure 6 shows the locations at which the cross-sections of the frame are extracted. The cross-sections are generated by generating planes at the two locations and then using them to slice the frame model.

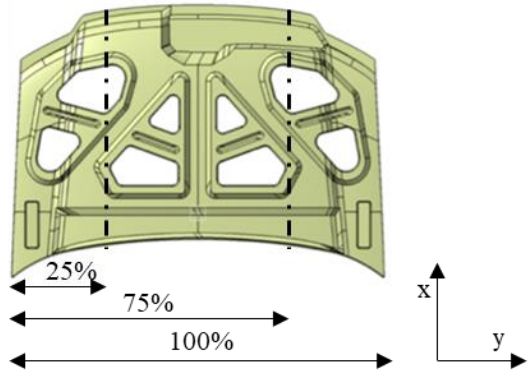

Figure 6: Frame with the 25% and 75% location for cross-sections

A custom Python script is used to generate the cross-sections from the geometry files in the dataset. The extracted cross-sections are recorded as $(x, z)$ points in CSV files – the representation type is textual data. However, due to variations in the size and shape of hood frames in the dataset, the number of points in a cross-section was not the same. Hence a standardized representation was established by defining a minimum required number of coordinates per cross-section to address this. The first and last 10% of the points were used to pad the boundary coordinates to preserve the overall shape and maintain a consistent length of the input data. This approach ensured consistency across the dataset while retaining the key geometric features of each cross-section. Figure 7 shows an example of the cross-section extracted from a hood frame.

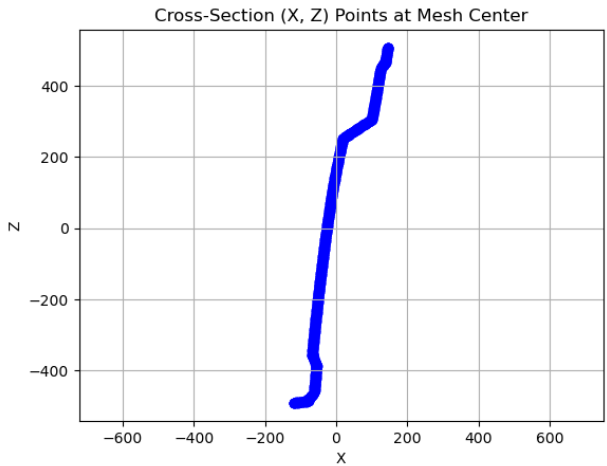

Figure 7: Example of cross-section points extracted from a hood frame

Although the cross-sections provide information on the variation in curvature of the hood frame, the ability to accurately capture *all* feature depths depends on the location of the slicing plane. In some cases, certain critical structural features may not intersect with the slicing plane which may impact the performance predictions. To address this limitation, the parametric values for depths of the primary features (or ribs) are extracted from the dataset using an automated script. This ensures that key geometric characteristics influencing structural performance are incorporated into the model, even if it is not captured in the cross-sections. This information is stored in CSV files.

A total of 20 design points from the final dataset of images, cross-sections, and parameters were saved separately for validating the model after training. These points were not used in the training or testing of the model. The rest of the data was split 80/20 for train and test datasets. The discussion in the results section utilizes the validation data to draw conclusions on the performance and efficiency of the model.

## 4. MODEL ARCHITECTURE

The research proposes to develop a multimodal neural network architecture to predict the performance metrics of car hood frames by integrating the three key data modalities: images, geometric data, and parametric data, described earlier. The proposed architecture consists of *five* individual networks and one final set of fully connected layers. Figure 8 shows the MMML architecture developed in this research. The *five* individual networks are:
1. Two separate convolutional networks (Residual Networks – ResNet50) to
   a. Extract feature maps from the top views
   b. Extract feature maps from the side views
2. Two neural networks to learn from the two cross-section datasets
3. One neural network to process the depth parameters in the frame model

Other networks like VGGNet [26], ResNet100, and AlexNet [27] were also tested and compared against ResNet50. The accuracy and computation resources, based on the dataset, remained unchanged. Hence the ResNet50 model was chosen as the network to process images in the MMML architecture. The input to each network is separate, and the networks don't interact with each other. Once all the relevant features are extracted from all the modalities, the results are concatenated and pooled into a unified representation. This is then fed into a network of fully connected layers to make the final prediction on the performance of the hood frame. By leveraging multimodal learning, this approach ensures a comprehensive understanding of design-performance relationships, enhancing prediction accuracy while reducing reliance on computationally expensive simulations.

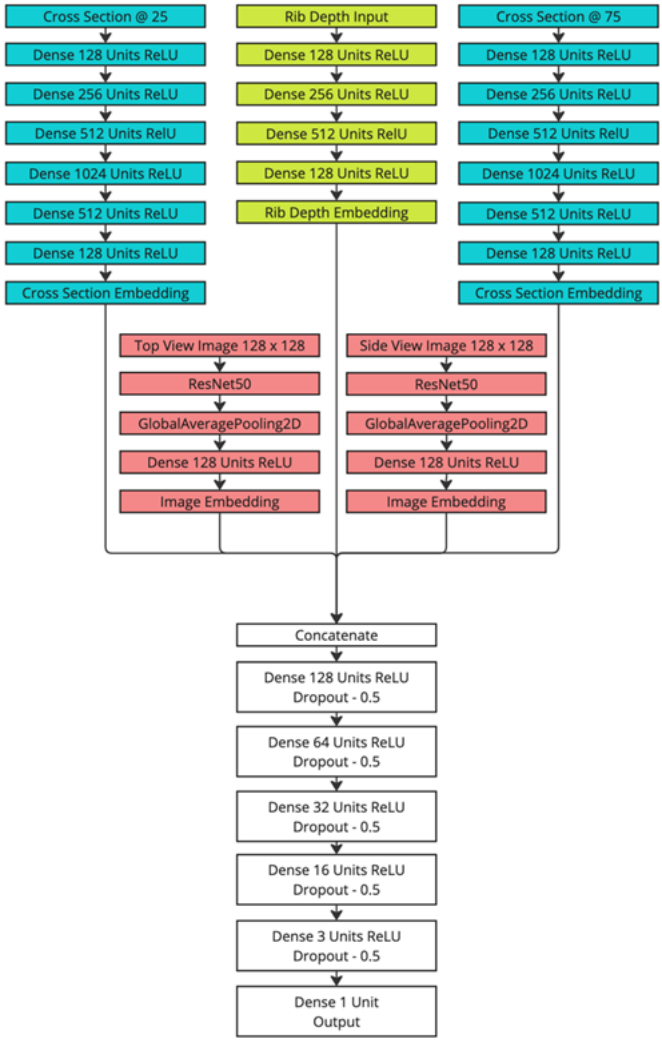

*Figure 8: Multi-Modal Machine Learning Architecture*

### 4.1 Creating Image Embeddings

The images are represented as *.jpg files of size 128 X 128 X 1 pixels - they are scaled and converted to grayscale. Given that a key characteristic of CNN models is extracting features, it is done by identifying various components of an image using filters. These filters, or vectors with assigned weights, are applied to an image to generate feature maps. The network attempts to learn and extract important patterns from the input data through this process. From a mathematical perspective, an image is represented as a matrix, where each cell corresponds to a pixel, and its value indicates the pixel's brightness on a scale from 0 to 1. Each pixel is assigned a weight, which can be referred to as its intensity. A filter, or a smaller matrix, is then applied to the image resulting in a feature map, which is a two-dimensional array that captures essential information about the features within the image. Figure 9 shows a sample of feature maps and a filter for a hood frame model in the dataset. Each feature map represents an attempt to identify a significant feature in the image. Given the dataset involves images of automotive hood frames, the CNN is expected to capture key geometric features, represented visually, such as the ribs, pockets, surface curvature, etc. For example, a top-view provides insight into how the hood features (primary and secondary ribs, pockets, hinges, and locks) are laid out and where the features on the hood are located. Similarly, a side-view shows the significance of the depth and curvature of the hood frame. These features will help the model better understand the shape and size of the features, along with their relative positions.

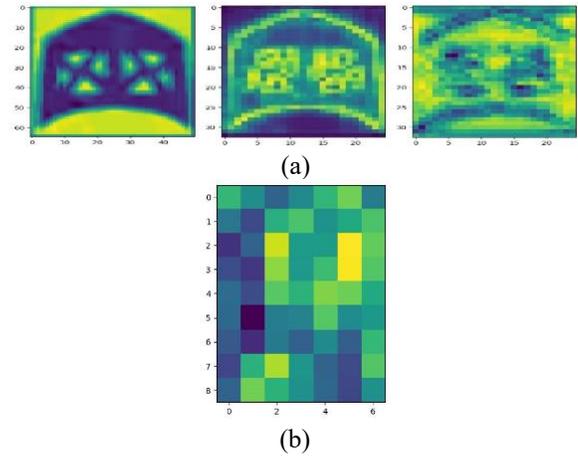

*Figure 9: Example visualization of (a) feature maps & (b) filter*

A ResNet50 base model is initialized without the "top layers", also known as the classification layers. The model is initialized without pre-trained weights; in other words, the weights are randomly assigned. This allows the model to learn directly from the dataset rather than relying on past datasets it has trained on. The extracted features are processed through a global pooling layer to reduce the spatial dimensions and produce a feature vector for each image. A fully connected layer (128 modes deep and a ReLU activation function) followed by batch normalization is used to capture complex nonlinear relationships within the data and to normalize activations to enhance training stability and convergence. As a result, the ResNet50 model outputs a 128-dimensional feature vector, which effectively encodes the most significant aspects of the image. Figure 10 shows the architecture of the two image-processing networks.

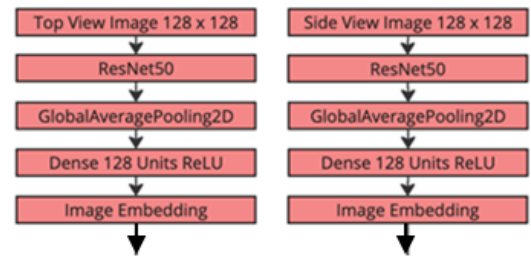

*Figure 10: ResNet50 architectures for top and side view images*

The feature vector from the CNN model would have captured all the important information from the images since, in a CNN model, with each subsequent convolution layer, the feature maps influence changes, and they focus on the finer details of the features as opposed to the whole image. These fine details of the features in the image would then be able to provide vital information to predict the performance of a hood frame.

## 4.2 Creating Parametric Data Embedding

Neural network models were created for processing the data associated with the two cross-sections and parametric data. The structure of the networks processing the cross-section data is shown in Figure 11. The input layer takes in the cross-section data, followed by a series of hidden layers. Each network has a total of six hidden layers of sizes 128, 256, 512, 1024, 512, and 128 nodes each, all utilizing the ReLU activation function.

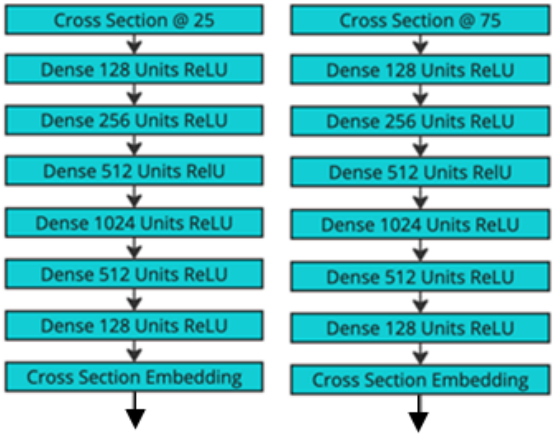

*Figure 11: Neural network architectures for cross-section data*

A Batch Normalization layer is applied to stabilize activations and improve training efficiency, functioning similarly to the batch normalization layer in the ResNet50 model. Finally, the output layer generates a 128-dimensional feature vector that embeds and captures the most critical features of the cross-sections.

The structure of the network processing the parametric data is shown in Figure 12. The input layer takes in the parameter (or depth) data, followed by a series of hidden layers. The network has a total of four hidden layers of sizes 128, 256, 512, and 128 nodes each, all utilizing the ReLU activation function. Similar to the networks processing the cross-sections, a batch normalization layer is used, which then generates a 128-dimensional vector that embeds the parametric data.

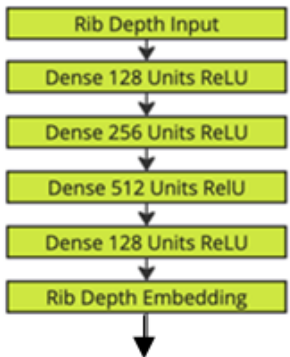

*Figure 12: Neural network for parametric data*

## 4.3 Performance Prediction

The fully connected layers play a crucial role in the final prediction. Once the significant features are extracted, the fully connected layers use these embeddings to make predictions on the output parameters. The fully connected layers thus serve as the decision-making component, interpreting various aspects of the input data and translating them into meaningful predictions regarding its performance.

In the context of hood frame prediction, the final component of the multimodal model is a neural network designed to predict the performance metrics of car hood frames. The feature embeddings from the ResNet50 models and the three neural networks are concatenated to create a unified representation. Figure 13 shows the architecture of the fully connected layers used to make the final prediction. The concatenated vector is processed through a series of hidden layers consisting of 128, 64, 32, 16, and 3 nodes each. The final dense layer includes three nodes, each responsible for predicting a distinct performance metric. L2 regularization is applied to prevent overfitting, ensuring that excessively large weight values are penalized during training. Additionally, dropout layers are utilized, with neurons being randomly deactivated at rates of 50%, 40%, and 30% across different layers. This regularization technique reduces reliance on specific neurons, thereby enhancing generalization. Finally, the output from the network consists of three units, each predicting a performance metric for the car hood frames.

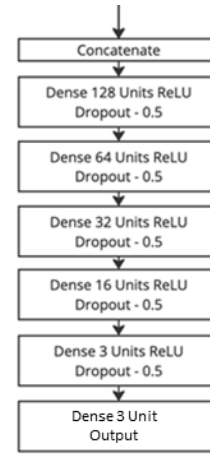

*Figure 13: Fully connected layers for final prediction*

## 5. RESULTS AND DISCUSSION

The most accurate predictions from a MMML model are achieved by fine-tuning various model parameters. Optimizing a model's architecture involves several key adjustments, such as modifying the number of layers (in both the initial and final models) and updating the number of nodes in the neural network. One of the main challenges in this process is striking the right balance between model complexity and simplicity. Finding the right model architecture and parameter required careful tuning. A model that was too simple would risk overfitting the training data, leading to poor generalization on the testing dataset. Conversely, an overly complex model could potentially struggle to extract meaningful insights from the data. The architecture described earlier in this paper represents the best balance tested so far—providing strong predictive performance without excessive complexity.

In addition to optimizing the model's architecture, fine-tuning the hyperparameters is equally critical in improving the performance of the model. Hyperparameters such as learning rate, batch size, patience value, and the number of training epochs all significantly impact the model's performance. The learning rate was one of the most important hyperparameters for the MMML model. The initial learning rate was set to 1e-5, which aids in rapid adjustments, but during training, as the number of epochs increases, both training and validation loss tend to plateau. This was countered using a dynamic function that updated (or lowered) the learning rate every time the loss plateaued. This dynamic nature of adjusting the learning rate enabled finer adjustments for improved performance and convergence.

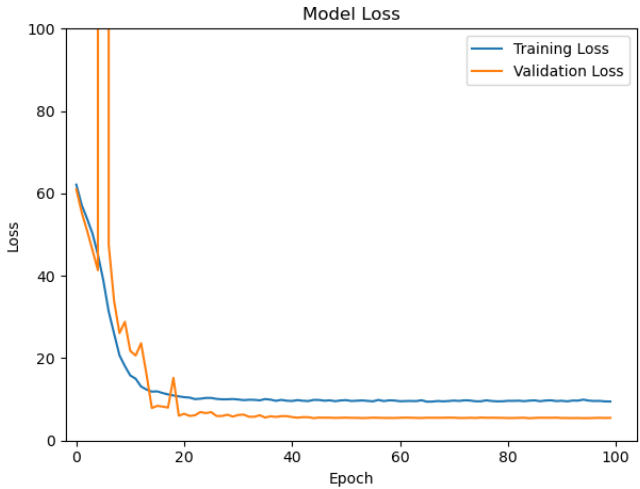

*Figure 14: Training and validation loss vs number of epochs*

The convergence behavior of the training and validation loss over 100 epochs is shown in Figure 14. The term validation loss in the figure is actually the "test" loss and not from the validation dataset. The validation dataset is used only after the training is complete. Initially, both losses exhibit a sharp decline, indicating rapid learning of key features from the data. A sudden spike in the validation loss at the beginning is primarily due to using random weights instead of pre-trained weights. Around the 20-epoch mark, the validation loss stabilizes, suggesting that the model begins refining its learned representations rather than simply memorizing patterns. Beyond 50 epochs, both losses approach an asymptotic limit, with minimal further reduction, demonstrating that the model has reached an optimal state. Notably, the validation loss remains closely aligned with the training loss throughout the training process. This indicates that the model is neither overfitting nor underfitting but generalizes well to unseen data. The absence of significant divergence between the two losses suggests that appropriate regularization techniques were applied, ensuring stable learning dynamics. The smooth convergence further highlights the benefits of an adaptive learning rate strategy, preventing premature stagnation or excessive oscillations in model performance. The model was trained on a Windows operating system, using an NVIDIA RTX A5000 GPU running on a 13th Gen Intel Core i9-13900K processor with 64GB RAM.

## 5.1 Results from Validation Dataset

Once training was complete, the model was validated using the validation dataset. Three separate graphs were generated to assess the model performance. Each plot shows the prediction of a performance metric – von Mises stress, mass, or deflection. The figures below illustrate the model's average error across these categories and individual error distributions for each data point.

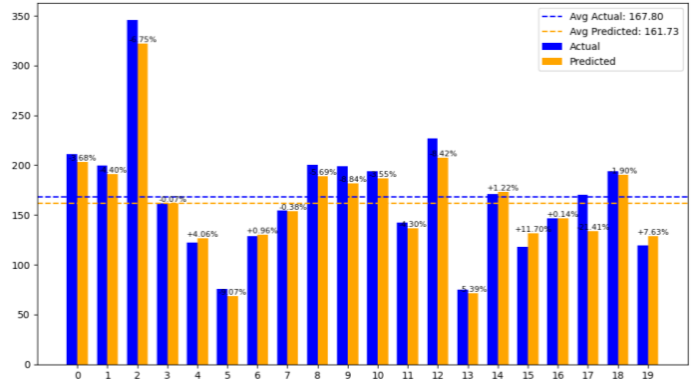

(a) vM stress (MPa) vs design points

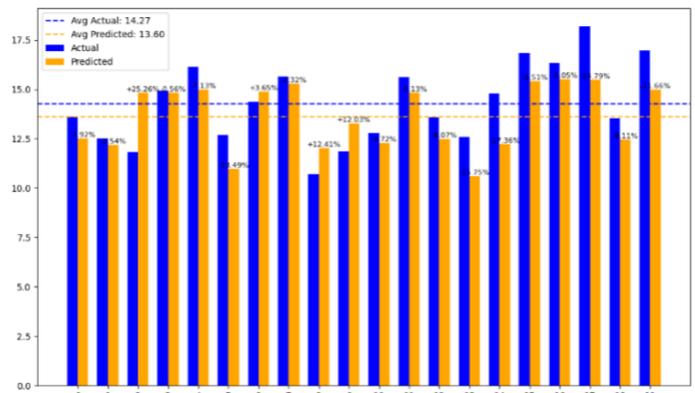

(b) Mass (kg) vs design points

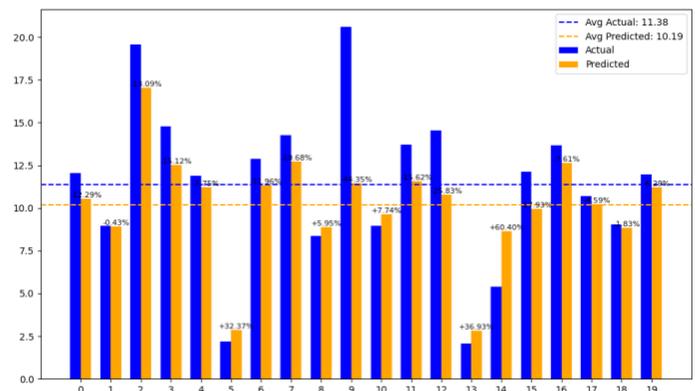

(c) Directional deflection (mm) vs design points

*Figure 15: Actual vs predicted values for the validation data (a) vM stress, (b) mass, and (c) directional deflection*

Table 1: Average error in the predicted values for the CarHoods10k dataset (based on Figure 15)

| Output Parameter | Error % |
|---|---|
| vM Stress | 5.48% |
| Mass | 9.27% |
| Deflection | 16.84% |

The MMML model performed quite well on the validation dataset, with error percentages well below 20%. These results were compared to another model using the same dataset and performance metrics as discussed by Kumar et al. in [18]. The results from this research are significantly better than the unimodal model (trained only on images) used in the other research. Kumar et al. had a confidence interval of 20% with a model accuracy of 90%, as compared to the MMML model, which had an accuracy of over 94% and confidence intervals in single digits for stress and mass and around 17% for deflection as noted in the table above.

### 5.2 Model Performance on Unseen Hood Frame Designs

To evaluate the generalizability of the proposed model, two hood frame designs, not part of CarHoods10k, were tested. The designs were sourced from an online community platform (GrabCAD [28]). Figure 16 and Figure 17 show the top and side views, along with the details on the cross-sections, of the two hood frames. These models underwent the same preprocessing steps, including data extraction and finite element analysis (FEA) using a similar model setup (to the CarHoods10k dataset). The FEA model setup for the two frames is shown in Figure 18. The FEA results served as ground truth, allowing the proposed model to compare the predicted performance values. Figure 19 shows the model's predictions of performance compared to the actual FEA results, and Table 2 shows the average error in prediction.

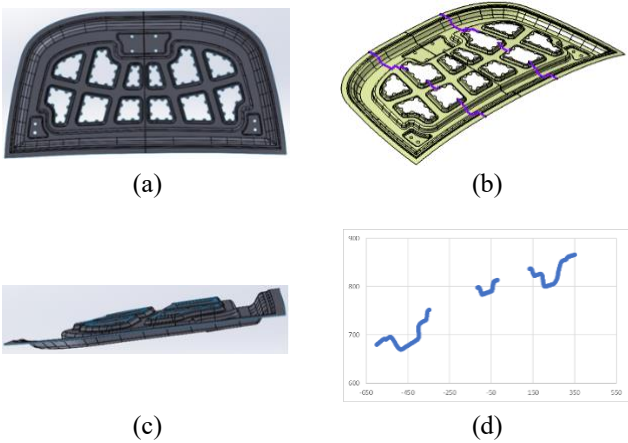

Figure 16: (a) Top view, (b) cross-section locations, (c) side view, and (d) cross-section points for Model 1

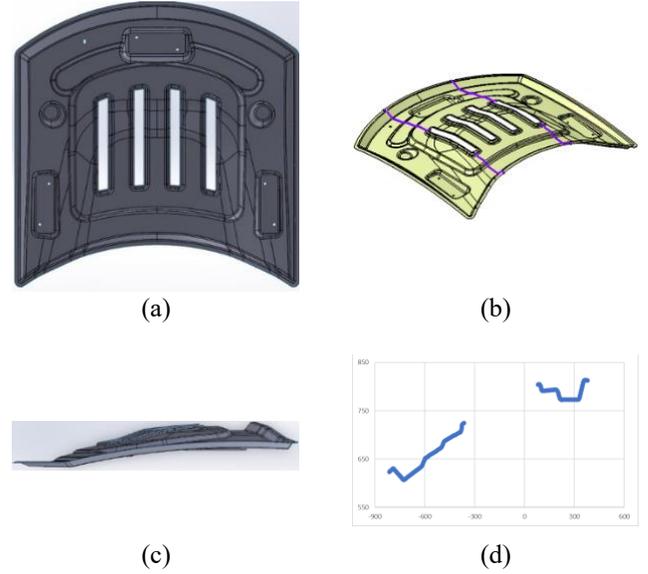

Figure 17: (a) Top view, (b) cross-section locations, (c) side view, and (d) cross-section points for Model 2

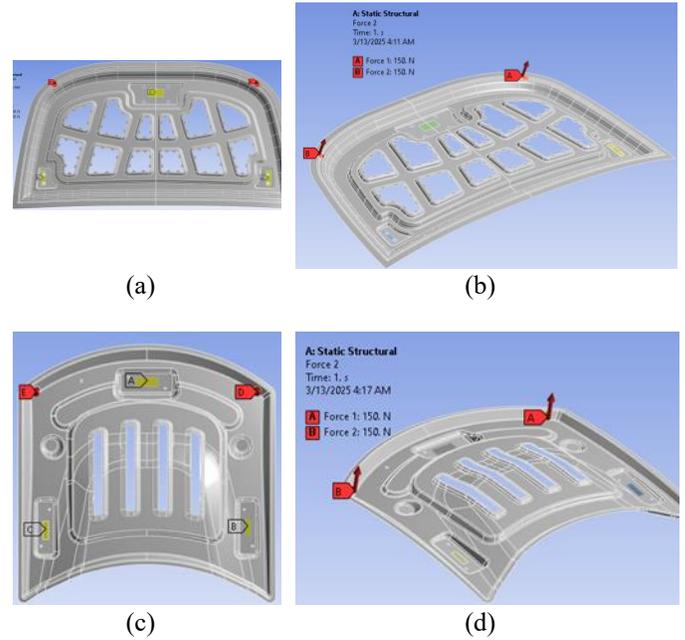

Figure 18: FEA setup for hood frames

As shown in Figure 19(a), the predicted stress values are slightly lower than the actual values (from FEA), with an average underprediction of 18.52%. While the model tends to be conservative in its stress estimates, the predictions are still within a reasonable range, capturing the general trend of stress distribution. This indicates that the model is not randomly over- or underestimating values but rather making consistent and physically plausible predictions, avoiding extreme overfitting to the training data.

Figure 19(b) shows that the model overpredicts mass, with an average error of 32.94%. While there is a tendency to overestimate mass, this does not indicate overfitting. Instead, the model slightly biases mass estimates towards safer, heavier designs. Given that mass estimation is influenced by multiple geometric parameters, small shifts in feature representation could explain these variations. Importantly, the model still captures the correct order of magnitude and does not diverge drastically from the actual value.

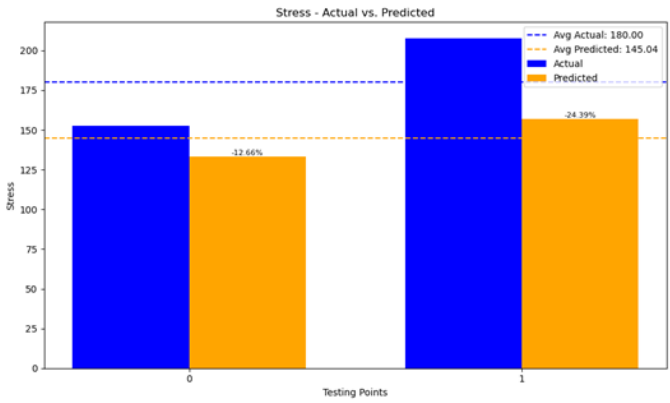

(a)

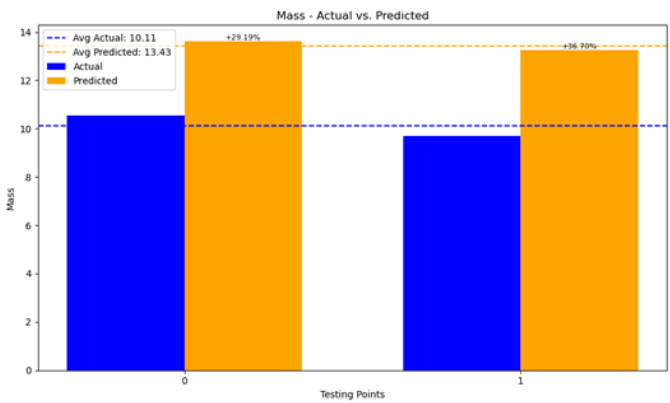

(b)

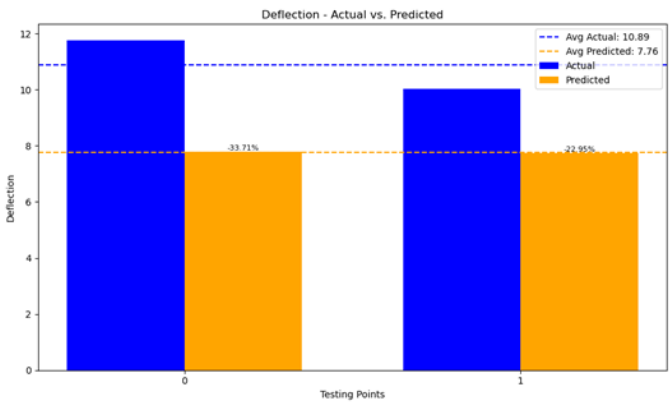

(c)

Figure 19: Model prediction vs actual values (a) stress, (b) mass, and (c) deflection

Figure 19(c) shows that the model's deflection predictions are lower than the FEA results with an average error of 28.33%. One potential reason for underprediction could be the presence of geometrical features (and their shapes) that are unique to this frame, and nothing remotely similar exists in the training dataset. However, the predictions remain physically reasonable and trend in the correct direction, reinforcing that the model is capturing the overall behavior of hood frame designs rather than memorizing specific patterns.

The results indicate a strong performance of the model, given that the new hood frames are not a part of the training dataset and the MMML model has never seen these before. This demonstrates the MMML model's robustness and ability to generalize beyond the training dataset.

Table 2: Actual vs predicted values (and average error) for new hood models

| Model \ Output | Hood Frame 1 | | Hood Frame 2 | | Avg. Error |
|---|---|---|---|---|---|
| | Actual Value | Predicted Value | Actual Value | Predicted Value | % |
| vM Stress (MPa) | 207.5 | 156.88 | 152.49 | 133.19 | 18.52% |
| Geometry Mass (kg) | 9.68 | 13.24 | 10.53 | 13.60 | 32.94% |
| Dir. Def. (mm) | 10.03 | 7.78 | 11.74 | 7.78 | 28.33% |

### 5.3 Comparing Unimodal vs Multimodal Performance

To quantitatively assess the effectiveness of the multimodal approach, the predictions from the MMML model were compared to predictions from a unimodal (image-based) architecture. Although the unimodal architecture only used images, it used a multiview-CNN approach [29], to consider both the top and side views in making the predictions. The evaluation was conducted for both the CarHoods10K datasets, as well as the two new hood frames.

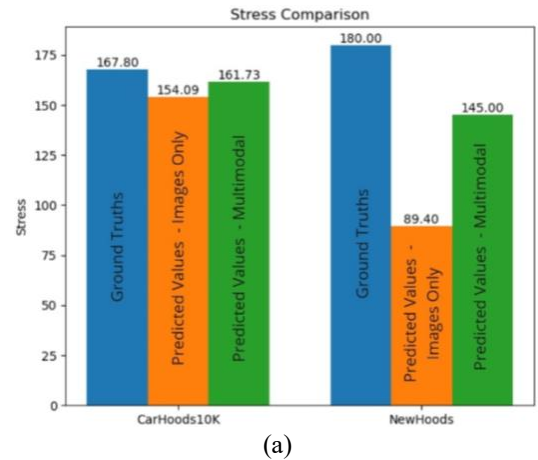

(a)

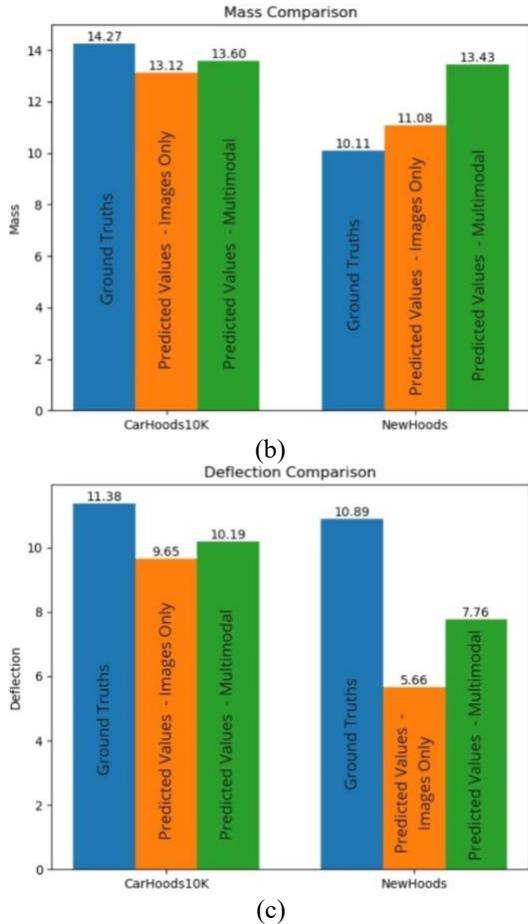

*Figure 20: Average model performance for unimodal (images only) vs multimodal for CarHoods10k and new hood frames*

The (average) prediction accuracies for the two models are compared to the actual values (from FEA) shown in Figure 20. The stress prediction results (Figure 20(a)) demonstrate that the multimodal model consistently outperforms the unimodal model by providing predictions closer to the actual values. The results are summarized in Table 3.

*Table 3: Summary of average error % in prediction for the two models on two different datasets*

| Dataset | Model | Stress Error % | Mass Error % | Def. Error % |
|---|---|---|---|---|
| CarHoods10k | Unimodal | 8.2% | 8.1% | 15.2% |
| | Multimodal | 3.6% | 4.7% | 10.5% |
| New Models | Unimodal | 50.3% | 9.6% | 48% |
| | Multimodal | 19.4% | 32.9% | 28.7% |

The results indicate that the MMML model consistently performs better on all the performance metrics on the CarHoods10k dataset. It also performs extremely well in improving the predictions of the stress and deflection values for the two unseen hood frame models. Overall, these findings demonstrate that multimodal models leveraging geometric, parametric, and image-based information outperform unimodal approaches, especially in out-of-distribution scenarios. The ability to maintain predictive accuracy on a previously unseen dataset supports the argument that multimodal learning is a robust framework for structural performance prediction tasks.

### 5.4 Summary

These results indicate that the multimodal learning model successfully generalizes to new designs without signs of overfitting (which would manifest as near-perfect predictions for training data but poor performance on unseen data) or severe underfitting (resulting in completely random or highly inaccurate estimates). The stress and deflection predictions are slightly conservative, which is preferable in many engineering applications where safety is a priority. While slightly overestimated, the mass predictions remain within a reasonable deviation and do not indicate an erratic or unstructured response. Importantly, the model correctly captures the relative differences between the two new designs, reinforcing that it has captured meaningful structural relationships.

### 6. CONCLUSION

By reducing reliance on computationally expensive simulations, the proposed MMML architecture enhances the efficiency of design exploration, enabling engineers to iterate rapidly while maintaining high-performance standards. This work contributes to the broader adoption of machine learning in engineering design, demonstrating its potential to supplement traditional simulation-based workflows and accelerate the development of optimized structures, especially in the conceptual design phase.

The proposed architecture also demonstrates that MMML outperforms single-modality approaches, highlighting the benefit of combining and using *different forms of the same data*. Additionally, the MMML model was tested on unseen frame models, showcasing its potential for generalization beyond the training dataset. The key contributions in the work can be summarized as:

- Developed an MMML architecture and demonstrated the use of multiple modalities to effectively train ML models
- Compared the performance of a multimodal architecture to an unimodal architecture and validated the effect of utilizing multiple modalities to yield better results
- Showcased the model's ability to extend beyond the CarHoods10K dataset, demonstrating its generalization to unseen data. However, additional testing on varied hood frame designs is necessary for broader applicability.

Ultimately, this study highlights the potential of MMML in reducing reliance on simulation, accelerating the design cycle, and enabling rapid performance evaluation. As machine learning continues to integrate into engineering workflows, refining multimodal approaches will be key to bridging the gap between predictive models and real-world applications.

### 6.1 Future Work

Despite these strengths, several areas for improvement remain. The current parametric representation, which relies on cross-sectional data, may not fully capture intricate geometric features such as ribs or pockets, which play a crucial role in structural performance. Future work will explore alternative parametric representations, potentially incorporating more granular feature-based data and geometric parameters to provide additional insight into the geometries. Similarly, the image input could be refined to focus on extracted features rather than full geometry images, allowing the model to learn more relevant patterns without redundant information. Future iterations of this model should investigate more structured feature selection techniques to preserve essential design characteristics while maintaining computational efficiency.

As the framework continues to evolve, balancing prediction accuracy with generalizability remains a priority. While reducing error rates is desirable, overly precise models risk overfitting, limiting their applicability to new and diverse designs. By continuously testing on significantly different hood frame geometries, future research will assess the robustness of the model across a broader range of engineering applications.

## ACKNOWLEDGEMENTS

The authors would like to thank Simon Vaglienti and Sandesh Kumawat for their contributions to the project.